\documentclass[11pt]{jmlr}
\makeatletter
\renewcommand{\ps@jmlrtps}{%
  \let\@mkboth\@gobbletwo
  \def\@oddhead{}\let\@evenhead\@oddhead
  \def\@oddfoot{\hfill\small\rmfamily\thepage\hfill}%
  \let\@evenfoot\@oddfoot}
\makeatother
\usepackage{times}
\usepackage{booktabs}
\newcommand{\E}{\mathbb E}
\newcommand{\Pp}{\mathbb P}
\newcommand{\ind}{\mathbf 1}
\newcommand{\cX}{\mathcal X}
\newcommand{\cY}{\mathcal Y}
\newcommand{\cF}{\mathcal F}
\newcommand{\cH}{\mathcal H}
\newcommand{\Risk}{\mathcal R}
\newcommand{\Opt}{\mathcal R^*}
\newcommand{\mass}{\mathsf M}
\title[Posterior Frequencies and the Cost of Not Fitting]{Posterior Frequencies and the Cost of Not Fitting:\\A Correction to a Memorization Lower Bound}
\author[Assylbekov and Takhanov]{%
\Name{Zhenisbek Assylbekov} \Email{zassylbe@pfw.edu}\\
\addr Department of Mathematical Sciences\\
Purdue University Fort Wayne
\AND
\Name{Rustem Takhanov} \Email{rustem.takhanov@nu.edu.kz}\\
\addr School of Computing and Artificial Intelligence\\
Nazarbayev University}
\hypersetup{pdfauthor={Zhenisbek Assylbekov and Rustem Takhanov},
 pdftitle={Posterior Frequencies and the Cost of Not Fitting: A Correction to a Memorization Lower Bound}}
\begin{document}
\maketitle
\begin{abstract}
Feldman's memorization lower bound assigns a frequency coefficient based on how often a point appears in the training sample. We give an exact three-point counterexample to its conditional and averaged forms, using independent fair labels and a permutation-equivariant learner. A sequence of sampling priors makes the ratio of the proposed penalty to actual excess risk grow exponentially, ruling out a universal positive constant-factor repair. The priors vary with sample size, the offending profiles become rare, and both quantities vanish. Retaining the original model, we obtain an exact replacement using full-sample posterior probabilities. Under exchangeability, these depend on multiplicity and the complete frequency profile. We characterize the largest multiplicity-only coefficients valid for unrestricted learners. The original coefficient has the posterior-mean form of Good's exact frequency formula and is an occupancy-weighted average of the corrected coefficients.
\end{abstract}

\section{Introduction}
If a learner misclassifies an observed noiseless label, it incurs population error whenever the corresponding input is encountered again. In a model with an unknown sampling distribution, the conditional cost is therefore the input's posterior probability given the training sample. \citet{feldman2021} uses this observation to relate fitting rare examples to generalization. His Theorem 2.3 assigns the same coefficient to every point observed $\ell$ times in a sample of size $n$:
\begin{equation}
 \tau_{n,\ell}=\frac{\E[Q^{\ell+1}(1-Q)^{n-\ell}]}{\E[Q^\ell(1-Q)^{n-\ell}]},
 \label{eq:tau-intro}
\end{equation}
where $Q$ is the random probability of a fixed domain point. This is the posterior mean of \(Q\), conditional only on the number of times the fixed point appears in the sample. It has the same functional form as \citet[formula (13), with $m=1$]{good1953}.

The learner, however, observes all sample counts. These may reveal additional information about $Q$. In Feldman's model, raw weights are drawn independently and then normalized; independence before normalization does not justify ignoring the other counts. 

Feldman's work provides an influential theoretical account of why learning from long-tailed distributions can require memorizing rare examples. Theorem~2.3 is a central step in that account: it connects errors on observed examples to excess population risk through frequency-dependent coefficients. Establishing the precise validity of this connection is therefore important beyond the particular conditioning identity used in its proof. We identify the failed conditioning step, give an exact counterexample to the theorem itself, and recover a valid lower bound within the original model.

The contributions are threefold. First, Section~\ref{sec:counterexample} gives an exact three-point counterexample, with independent fair labels and a permutation-equivariant learner. Second, Section~\ref{sec:quantitative} shows that the failure can be exponentially large in relative terms: along even sample sizes $n$, a sequence of three-entry frequency lists gives
\[
 \frac{\text{proposed averaged penalty}}{\text{actual excess risk}}
 \sim\frac{2}{15}\left(\frac43\right)^n.
\]
Thus no universal positive constant multiple of the original penalty is valid. The prior changes with $n$, the learner errs on an increasingly rare count profile, and both the penalty and excess risk tend to zero. This is a worst-case multiplicative obstruction, not a fixed-prior asymptotic result or a nonvanishing additive gap.

Third, Sections~\ref{sec:corrected}--\ref{sec:good} give the exact replacement and its sharp multiplicity-only relaxation. Under a permutation-invariant prior, the full-sample posterior depends only on the point's multiplicity and the frequency profile: the numbers of domain points appearing zero times, once, twice, and so forth. A multiplicity-only coefficient is valid for all learners exactly when it does not exceed the smallest feasible profile coefficient. Feldman's \(\tau_{n,\ell}\) instead averages these coefficients with occupancy weights, allowing learners to select errors on profiles where the average is too large.

\paragraph{Related work and scope.}
The posterior decomposition and exchangeability arguments are standard; their role here is to identify the exact correction. The main contributions are the counterexample, its exponential relative amplification, and the sharp criterion for universal learner-dependent bounds. Count-only prediction is closely related to Johnson's sufficientness postulate: under suitable assumptions across sample sizes, it leads to classical Dirichlet characterizations \citep{zabell1982}. Our criterion concerns one fixed sample size and observed categories; it is not a characterization of Dirichlet priors. Section~\ref{sec:sufficientness} makes this distinction precise. \citet{lee2013} characterize when predictive rules based on species counts are compatible with an exchangeable sampling model. Our question is different: given a sampling model, which posterior probabilities yield valid lower bounds on a learner's excess risk? Appendix~\ref{sec:integral} gives a finite-dimensional Gamma-integral calculation related to posterior analysis of normalized random measures \citep{james2009}; it is a computational aid, not a new posterior method.

We correct Theorem 2.3 of arXiv:1906.05271v4, dated January 10, 2021, while retaining its normalized-weight model. Section~\ref{sec:dependencies} identifies specific downstream dependencies and separates them from calculations that do not use the failed conditioning identity. Fixed-prior and typical-profile approximations, including long-tail guarantees, require additional analysis.

\section{Model and posterior probabilities}\label{sec:model}
Let $\cX=[N]=\{1,\ldots,N\}$ be a finite domain and let $\cY$ be a finite label set with at least two elements. A prior $\Pi$ is a probability distribution on the simplex
\[
 \Delta_N=\{d\in[0,1]^N:\textstyle\sum_xd_x=1\}.
\]
Independently draw a probability vector $D\sim\Pi$ and a labeling function $f\sim\cF$, where $\cF$ is any distribution on $\cY^{\cX}$. Independence here is between $D$ and $f$; the labels $f(x)$ may be dependent across points. Conditional on $(D,f)$, draw $U=(X_1,\ldots,X_n)\sim D^n$ and observe
\[
 S=((X_1,f(X_1)),\ldots,(X_n,f(X_n))).
\]
All claims conditional on a dataset concern datasets of positive probability. We assume $n\ge1$. A learner $A$ maps each dataset to a distribution on $\cY^{\cX}$; its randomization is independent of the latent variables conditional on the dataset. No computational or hypothesis-class restriction is imposed.

For a realized sample $Z$, define the counts and the observed set by
\[
 c_x=c_x(Z)=\sum_{t=1}^n\ind\{X_t=x\},\qquad
 O(Z)=\{x:c_x>0\}.
\]
Noiselessness makes the label $y_Z(x)$ well-defined for every $x\in O(Z)$. Let
\[
 L_c(d)=\prod_{x\in\cX}d_x^{c_x},\qquad 0^0:=1.
\]
Define \(q_x(Z)\) as the posterior mean of \(D_x\). As shown in Lemma~\ref{lem:factor} below, it satisfies
\begin{equation}
 q_x(Z):=\E[D_x\mid S=Z]
 =\frac{\E_{D\sim\Pi}[D_xL_c(D)]}
 {\E_{D\sim\Pi}[L_c(D)]}.
 \label{eq:posterior}
\end{equation}
We also write $q_x(c)$ because this quantity depends only on the input counts.

\begin{lemma}[Posterior factorization]\label{lem:factor}
Conditional on $S=Z$, the random variables $D$ and $f$ are independent. The posterior of $D$ depends only on $c$, and the posterior of $f$ is its prior conditioned on agreeing with all observed labels. In particular, \eqref{eq:posterior} holds.
\end{lemma}
\begin{proof}
For a fixed ordered dataset, its likelihood given $(D,f)$ is
\[
 L_c(D)\,\ind\{f(x)=y_Z(x)\text{ for every }x\in O(Z)\}.
\]
The two factors involve separate variables, and the joint prior is a product. Normalization therefore yields a product posterior. Integrating $D_x$ against its factor gives \eqref{eq:posterior}.
\end{proof}

\paragraph{Feldman's sampling prior.}
Given a list $\pi=(\pi_1,\ldots,\pi_N)$ of positive numbers, independently choose each $W_x$ uniformly from its entries and set
\begin{equation}
 D_x=\frac{W_x}{T},\qquad T=\sum_{z=1}^NW_z.
 \label{eq:normalized}
\end{equation}
Repeated entries are counted with their multiplicities. The list may sum to one, but the independently sampled weights need not. More generally, one may use any common distribution on positive raw weights. This construction gives an exchangeable prior: permuting the coordinates of $D$ does not change its distribution. We write $\bar\pi^N$ for the marginal law of $D_x$ in the list model, following Feldman. This law generally differs from the uniform law on the raw list.

\paragraph{Count-only conditioning.}
For a fixed $x$, conditional on $D_x=\alpha$, the count $C_x=c_x(S)$ has distribution $\operatorname{Binomial}(n,\alpha)$. Consequently,
\begin{equation}
 \E[D_x\mid C_x=\ell]
 =\frac{\E[D_x^{\ell+1}(1-D_x)^{n-\ell}]}
 {\E[D_x^\ell(1-D_x)^{n-\ell}]}
 =\tau_{n,\ell}(x),
 \label{eq:count-posterior}
\end{equation}
whenever the denominator is positive. Under an exchangeable prior the value does not depend on $x$, and we write $\tau_{n,\ell}$ for this common value. Equation~\eqref{eq:count-posterior} is valid; the assertion $q_x(Z)=\tau_{n,c_x}$ for every dataset is the additional claim that fails.

\paragraph{Risk and error-count conventions.}
For a dataset $Z$, let
\[
 \Risk(A\mid Z)=\E_{D,f\mid Z}\E_{h\sim A(Z)}
 \left[\sum_xD_x\ind\{h(x)\ne f(x)\}\right],\qquad
 \Opt(Z)=\inf_B\Risk(B\mid Z).
\]
The infimum is over all learners. Thus $\Opt(Z)$ is the minimum posterior risk achievable from the same observed data $Z$ and the specified priors; the predictor is not given the latent labeling function or sampling distribution. Write $\Risk(A)=\E[\Risk(A\mid S)]$ and $\Opt=\inf_B\Risk(B)$ for the corresponding averaged quantities. For observed points define
\[
 e_x(A,Z)=\Pp_{h\sim A(Z)}\{h(x)\ne y_Z(x)\},\qquad
 B_\ell(A,Z)=\sum_{x:c_x=\ell}e_x(A,Z),\quad \ell\ge1.
\]
Thus $B_\ell$ counts misclassified distinct points in expectation over the learner's randomness, with no division by $n$ and no extra factor $\ell$. It is Feldman's $\operatorname{errn}_Z(A,\ell)$. In this notation the original assertion is
\begin{equation}
 \Risk(A\mid Z)-\Opt(Z)\ge\sum_{\ell=1}^n\tau_{n,\ell}B_\ell(A,Z),
 \label{eq:original-bound}
\end{equation}
for every learner and dataset, together with its averaged version. We now show that both assertions can fail.

\section{An exact counterexample}\label{sec:counterexample}
We use $N=n=3$ and the raw list
\[
 \pi=(1/4,1/4,1/2).
\]
Thus $W_i=1/4$ with probability $2/3$ and $W_i=1/2$ with probability $1/3$, independently for $i=1,2,3$. Direct averaging over the $27$ equally likely choices of list entries gives
\begin{align}
 \tau_{3,1}
 &=\frac{\E[D_1^2(1-D_1)^2]}{\E[D_1(1-D_1)^2]}
 =\frac{226313}{682260}\approx0.331711,
 \label{eq:counter-tau}\\
 \E[D_1\mid U=(1,2,2)]
 &=\frac{\E[D_1^2D_2^2]}{\E[D_1D_2^2]}
 =\frac{118273}{358140}\approx0.330242.
 \label{eq:counter-post}
\end{align}
In contrast, exchangeability and $\sum_iD_i=1$ imply
\begin{equation}
 \E[D_1\mid U=(1,2,3)]=1/3.
 \label{eq:counter-distinct}
\end{equation}
The first point occurs once in both sequences, but its posterior means differ. Appendix~\ref{app:certificate} supplies an exact-arithmetic certificate.

\begin{proposition}[Failure with independent fair labels]\label{prop:failure}
Let $N=n=3$, and let the sampling prior $\Pi$ be the law of $D_i=W_i/(W_1+W_2+W_3)$, where the $W_i$ are independent, with $\Pp(W_i=1/4)=2/3$ and $\Pp(W_i=1/2)=1/3$. Independently of $D$, draw the labels $f(1),f(2),f(3)$ independently from $\operatorname{Bernoulli}(1/2)$: each label is $0$ or $1$ with equal probability. Each point's label is drawn once and is shared by all occurrences of that point in the sample. For these sampling and labeling priors, there exists a permutation-equivariant learner for which both \eqref{eq:original-bound} and its averaged version fail.
\end{proposition}
\begin{proof}
Consider the event $E$ that the counts are a permutation of $(1,2,0)$. The learner predicts all observed labels correctly except that, on $E$, it flips the label of the unique singleton. On each unseen point it predicts $0$ or $1$ with equal probability, using independent random bits. This rule is equivariant under permutations of the domain.

The learner is posterior-optimal on every unseen point. Thus, conditional on a sample in $E$, its excess risk is exactly the posterior mass of the singleton, namely \eqref{eq:counter-post}. Its only observed-point error is that singleton, so the proposed penalty is \eqref{eq:counter-tau}. This is a strict violation. Outside $E$ its excess risk and penalty are both zero. By exchangeability, averaging multiplies both quantities by the same positive factor $\Pp(E)$ and preserves the violation. For example, $\Pp(E)=6\cdot3\E[D_1D_2^2]=5969/9000$.
\end{proof}

A constant-zero labeling prior gives an even shorter witness: use the same rule with zero predictions except for the singleton flip. Then the optimal risk is zero. The fair-label construction shows that the failure does not depend on knowing all labels in advance.

\paragraph{The conditioning error.}
The proof of Lemma~A.1 in  \citet[v4]{feldman2021} asserts that, given $D_x=\alpha$, the relative weights of all other points have the law of fresh independent raw weights normalized over those points. For the three-point construction with raw list $\pi=(1/4,1/4,1/2)$, the identity $D_1=W_1/(W_1+W_2+W_3)$ gives
\(D_1=1/2\,\,\Longleftrightarrow\,\, W_1=W_2+W_3\). 
Since each $W_i$ belongs to $\{1/4,1/2\}$, we have $W_1\le1/2$ and $W_2+W_3\ge1/2$. Equality therefore forces
\(
 (W_1,W_2,W_3)=(1/2,1/4,1/4)
\). 
Hence $D_2/(1-D_1)=D_3/(1-D_1)=1/2$ deterministically under that conditioning. In contrast, two fresh independent draws from the raw list can be $1/4$ and $1/2$, yielding relative weights $1/3$ and $2/3$ after normalization. The asserted conditional construction is therefore invalid. The counterexample uses the original finite-list model exactly; no limit, approximation, or alternative prior is involved.

\section{An exponentially large relative failure}\label{sec:quantitative}
The three-point counterexample has a small numerical discrepancy. We now show that this is not a uniform limitation: even with $N=3$, the original coefficient can exceed the correct one by an arbitrarily large factor.

For even $n\ge4$, put $\varepsilon_n=2^{-n/2}$ and use the probability list
\begin{equation}
 \pi^{(n)}=\frac{1}{1+2\varepsilon_n}
       (\varepsilon_n,\varepsilon_n,1).
 \label{eq:family}
\end{equation}
Multiplying every raw weight by the same constant leaves the normalized vector unchanged. We may therefore compute with weights equal to $\varepsilon_n$ with probability $2/3$ and to $1$ with probability $1/3$. Let $\Pi_n$ denote the resulting prior on $D$. Let $E_n$ be the event that the count vector is a permutation of $(1,n-1,0)$, and define the singleton's posterior mass on this event by
\[
 q_{n,1}=\E_{\Pi_n}[D_1\mid (C_1,C_2,C_3)=(1,n-1,0)].
\]
Here $1$ denotes a singleton; the profile $(1,n-1,0)$ is implicit. By exchangeability, this is the singleton's posterior mass for every sample in $E_n$. Throughout this section, $\tau_{n,\ell}$ is computed under the sample-size-dependent prior $\Pi_n$; the same convention applies in Appendix~\ref{app:limits}.
The notation $a_n\sim b_n$ means $a_n/b_n\to1$, here along even sample sizes.

\begin{theorem}[No universal constant-factor repair]\label{thm:exponential}
For the priors \eqref{eq:family},
\begin{equation}
 q_{n,1}\sim\frac54\,2^{-n/2},\qquad
 \tau_{n,1}\sim\frac16\left(\frac{2\sqrt2}{3}\right)^n,
 \qquad
 \frac{\tau_{n,1}}{q_{n,1}}\sim\frac{2}{15}\left(\frac43\right)^n.
 \label{eq:exponential}
\end{equation}
With independent fair binary labels there is a permutation-equivariant learner $A_n$ whose proposed averaged penalty
$P_n=\sum_{\ell=1}^n\tau_{n,\ell}\E[B_\ell(A_n,S)]$ and actual excess risk satisfy
\begin{equation}
 \frac{P_n}{\Risk(A_n)-\Opt}=\frac{\tau_{n,1}}{q_{n,1}}\longrightarrow\infty.
 \label{eq:ratio-risk}
\end{equation}
Consequently, no constant $c>0$, independent of the frequency list and sample size, makes $\Risk(A)-\Opt\ge c\sum_\ell\tau_{n,\ell}\E[B_\ell(A,S)]$ valid for all such models and learners.
\end{theorem}
\begin{proof}
Write $\varepsilon=\varepsilon_n$ and define the four expectations
\begin{align*}
 a_n&=\E[D_1(1-D_1)^{n-1}],&
 b_n&=\E[D_1^2(1-D_1)^{n-1}],\\
 z_n&=\E[D_1D_2^{n-1}],&
 v_n&=\E[D_1^2D_2^{n-1}].
\end{align*}
Thus $\tau_{n,1}=b_n/a_n$ and $q_{n,1}=v_n/z_n$. The singleton count alone leaves posterior weight on all-equal raw configurations. Observing that another point occurs $n-1$ times assigns those configurations a much smaller likelihood. Quantitatively, the eight raw-weight configurations give
\begin{equation}
 a_n\sim\frac{\varepsilon}{3},\qquad
 b_n\sim\frac1{27}\left(\frac23\right)^{n-1},\qquad
 z_n\sim\frac4{27}\varepsilon,\qquad
 v_n\sim\frac5{27}\varepsilon^2.
 \label{eq:four-limits}
\end{equation}
Here is the leading-term calculation. Since $n\varepsilon\to0$, factors of the form $(1+O(\varepsilon))^{n}$ tend to one. For $a_n$, the configurations $(\varepsilon,1,\varepsilon)$ and $(\varepsilon,\varepsilon,1)$ each contribute asymptotically $(4/27)\varepsilon$, while $(\varepsilon,1,1)$ contributes asymptotically $\varepsilon/27$. The all-equal configurations have total probability $1/3$ and normalized vector $(1/3,1/3,1/3)$; they supply the leading term of $b_n$. Their contribution to $a_n$ is negligible because $(2/3)^n=o(\varepsilon)$. All other contributions to $b_n$ are negligible because $\varepsilon^2=2^{-n}=o((2/3)^n)$, or decay faster.

For $z_n$, only $(\varepsilon,1,\varepsilon)$ contributes at order $\varepsilon$. For $v_n$ this configuration contributes asymptotically $(4/27)\varepsilon^2$, while $(1,1,\varepsilon)$ contributes
\[
 \frac2{27}(2+\varepsilon)^{-(n+1)}
 \sim\frac1{27}2^{-n}=\frac1{27}\varepsilon^2.
\]
The other terms are smaller. Appendix~\ref{app:limits} lists every configuration's scaled limit. Taking ratios gives \eqref{eq:exponential}.

The learner $A_n$ predicts observed labels correctly except that it flips the unique singleton on $E_n$; it predicts independent fair bits on unseen points. This is permutation-equivariant and optimal except at the flipped point. Hence its excess risk is $\Pp(E_n)q_{n,1}$, whereas $P_n=\Pp(E_n)\tau_{n,1}$. The event has positive probability, proving \eqref{eq:ratio-risk} and the final assertion.
\end{proof}

\paragraph{Size of the event and absolute risks.}
There are six assignments of the counts $(1,n-1,0)$ to named points and $n$ orders for each assignment. Thus
\begin{equation}
 \Pp(E_n)=6nz_n\sim\frac89 n2^{-n/2},
 \label{eq:event-size}
\end{equation}
and
\begin{equation}
 \Risk(A_n)-\Opt\sim\frac{10}{9}n2^{-n},\qquad
 P_n\sim\frac4{27}n\left(\frac23\right)^n.
 \label{eq:absolute-size}
\end{equation}
Both quantities vanish. The theorem excludes a uniform multiplicative repair over frequency lists, including on a fixed three-point domain. It does not exclude useful additive-error bounds, bounds on typical profiles, or approximations for a fixed long-tailed prior. The list in \eqref{eq:family} depends on $n$, which is essential to the scope of this conclusion. For $n=32$, exact enumeration gives $\tau_{32,1}/q_{32,1}\approx1234.9$; Appendix~\ref{app:limits} provides further finite-sample values.

\section{The corrected bound and its profile reduction}\label{sec:corrected}
For the learner and label posterior, define

\[
 a_{x,y}(Z)=\Pp_{h\sim A(Z)}(h(x)=y),\qquad
 \eta_{x,y}(Z)=\Pp(f(x)=y\mid S=Z).
\]
The posterior label regret at $x$ is
\begin{equation}
 r_x(A,Z)=\max_{y\in\cY}\eta_{x,y}(Z)
       -\sum_{y\in\cY}a_{x,y}(Z)\eta_{x,y}(Z)\ge0.
 \label{eq:regret}
\end{equation}
By noiselessness, $r_x(A,Z)=e_x(A,Z)$ when $c_x>0$.

\begin{theorem}[Exact posterior decomposition]\label{thm:decomposition}
For every sampling prior $\Pi$, independent labeling prior $\cF$, learner $A$, and positive-probability dataset $Z$,
\begin{equation}
 \Opt(Z)=\sum_{x\in\cX}q_x(Z)
       \left(1-\max_{y\in\cY}\eta_{x,y}(Z)\right),
 \label{eq:opt}
\end{equation}
and
\begin{equation}
 \Risk(A\mid Z)-\Opt(Z)
 =\sum_{x:c_x>0}q_x(Z)e_x(A,Z)
  +\sum_{x:c_x=0}q_x(Z)r_x(A,Z).
 \label{eq:decomp}
\end{equation}
In particular, dropping the second sum gives a lower bound. Equality in that lower bound is achieved by every learner that makes posterior-optimal predictions on unseen points.
\end{theorem}
\begin{proof}
By Lemma~\ref{lem:factor} and the independence of the learner's randomization,
\[
 \Risk(A\mid Z)
 =\sum_x q_x(Z)
       \left(1-\sum_y a_{x,y}(Z)\eta_{x,y}(Z)\right).
\]
Because the output class contains all functions on $\cX$, the conditional risk can be minimized separately at each point. Choosing any maximizing label for each $\eta_{x,y}$ gives \eqref{eq:opt}. Subtracting yields $\sum_xq_xr_x$. For observed points the label posterior is a point mass at $y_Z(x)$, giving \eqref{eq:decomp}. A maximizing label on each unseen point sets its regret to zero.
\end{proof}

An alternative proof of the inequality modifies $A$ by correcting all its observed-point predictions and keeping its other predictions. The decrease in risk is exactly the first sum in \eqref{eq:decomp}. The modified learner cannot have risk below $\Opt(Z)$. This proof explains the cost of not fitting without any exchangeability assumption.

\paragraph{Sharpness.}
For any chosen observed point $x$, a learner can make a deliberate error only at $x$ and use optimal predictions everywhere else. Its excess risk is exactly $q_x(Z)$. Thus the individual weights in the lower bound cannot be increased while retaining a statement for all learners. A point observed with positive count necessarily has positive posterior mass under a positive-probability dataset.

Since the data space is finite and pointwise minimizers exist,
\[
 \Opt:=\inf_B\Risk(B)=\E[\Opt(S)].
\]
Averaging \eqref{eq:decomp} consequently gives a valid unconditional decomposition as well. The posterior coefficient remains inside the expectation; it cannot in general be replaced by a count-only mean.

\subsection{Reduction to the frequency profile}\label{sec:profile}
Assume from now on that $\Pi$ is exchangeable. Define the profile
\[
 H=(K_0,\ldots,K_n),\qquad
 K_j=\#\{x:C_x=j\}.
\]
Every profile satisfies $\sum_jK_j=N$ and $\sum_jjK_j=n$. The coordinate $K_0$ records the number of unseen domain points; $N$ is known in this model. A profile is \emph{feasible} if it has positive probability. Let $\cH_{N,n}$ denote the finite set of feasible profiles.

\begin{lemma}[Profile reduction]\label{lem:profile}
There are functions $\theta_{n,\ell}(h)$, defined whenever $h\in\cH_{N,n}$ and $h_\ell>0$, such that
\[
 q_x(Z)=\theta_{n,c_x}(H(Z)).
\]
In particular, all points of the same multiplicity have the same posterior mean, regardless of their observed labels.
\end{lemma}
\begin{proof}
If two count vectors have the same profile, a permutation of domain points maps one to the other. If their designated coordinates have the same count, the permutation can be chosen to map those coordinates to each other. Apply this change of variables to numerator and denominator of \eqref{eq:posterior}; exchangeability preserves the prior. Lemma~\ref{lem:factor} removes the dependence on labels.
\end{proof}

\begin{corollary}[Corrected profile-dependent bound]\label{cor:profile}
Under an exchangeable sampling prior,
\begin{equation}
 \Risk(A\mid Z)-\Opt(Z)
 \ge\sum_{\substack{\ell\ge1:\ K_\ell(Z)>0}}
       \theta_{n,\ell}(H(Z))B_\ell(A,Z).
 \label{eq:profile-bound}
\end{equation}
Its averaged form is
\begin{equation}
 \Risk(A)-\Opt
 \ge\E\left[\sum_{\substack{\ell\ge1:\ K_\ell>0}}
       \theta_{n,\ell}(H)B_\ell(A,S)\right].
 \label{eq:profile-average}
\end{equation}
Both bounds are attained by learners that predict optimally on unseen points, with arbitrary predictions on observed points.
\end{corollary}
\begin{proof}
Substitute Lemma~\ref{lem:profile} into Theorem~\ref{thm:decomposition}, group the observed points by multiplicity, and then average.
\end{proof}

The sampling prior must be exchangeable, but the labeling prior and the learner need not be. In particular, label structure does not destroy the reduction of the frequency posterior to the profile. It affects the unseen-point regret and the learner's pattern of errors.

The corrected coefficients obey the normalization identity
\begin{equation}
 \sum_{\ell:K_\ell>0}K_\ell\theta_{n,\ell}(H)=1,
 \label{eq:normalization}
\end{equation}
including $\ell=0$. Computing them requires the sampling prior; the profile alone does not specify a prior-free numerical value.

\section{Good's formula and sharp count-only bounds}\label{sec:good}
\subsection{An exact averaging identity}
Define the random total probability of the points with multiplicity $\ell$ by
\[
 \mass_\ell=\sum_xD_x\ind\{C_x=\ell\}.
\]
This quantity depends on both $D$ and the sample. Conditional on the profile, its expectation is
\begin{equation}
 \E[\mass_\ell\mid H]
 =K_\ell\theta_{n,\ell}(H),
 \label{eq:mass-profile}
\end{equation}
where the product is interpreted as zero when $K_\ell=0$.

\begin{proposition}[Relation to the count-only coefficient]\label{prop:average}
For every $\ell$ with $\Pp(C_x=\ell)>0$,
\begin{equation}
 \tau_{n,\ell}
 =\E[\theta_{n,\ell}(H)\mid C_x=\ell]
 =\frac{\E[K_\ell\theta_{n,\ell}(H)]}{\E[K_\ell]}
 =\frac{\E[\mass_\ell]}{\E[K_\ell]}.
 \label{eq:averaging}
\end{equation}
\end{proposition}
\begin{proof}
Apply the tower property to $\E[D_x\mid C_x=\ell]$ and use Lemma~\ref{lem:profile}. Alternatively, exchangeability gives
\[
 \E[\mass_\ell]=N\E[D_x\ind\{C_x=\ell\}],\qquad
 \E[K_\ell]=N\Pp(C_x=\ell).
\]
Their ratio is \eqref{eq:count-posterior}; \eqref{eq:mass-profile} gives the middle expression.
\end{proof}

Thus the averaging is weighted by the number of points in the multiplicity class. An unweighted expectation over profiles is generally different. A profile containing more singletons is more likely under conditioning on a specified point being a singleton.

For Good's fixed-population model, let $p_1,\ldots,p_N$ be fixed population probabilities and choose an index $I$ uniformly from $[N]$, independently of the sample. Then
\begin{equation}
 \E[p_I\mid C_I=\ell]
 =\frac{\sum_i p_i^{\ell+1}(1-p_i)^{n-\ell}}
        {\sum_i p_i^\ell(1-p_i)^{n-\ell}}.
 \label{eq:good}
\end{equation}
This is formula (13), with $m=1$, of \citet{good1953}, after changing notation. It has the same functional form as \eqref{eq:count-posterior}. Good averages over a uniformly selected species in a fixed population; Feldman's coefficient averages over the marginal law of a normalized random frequency.

One may also put a uniform random permutation prior on the fixed vector $(p_1,\ldots,p_N)$. The marginal frequency law is then $N^{-1}\sum_i\delta_{p_i}$, where $\delta_p$ is a unit point mass at $p$, so \eqref{eq:good} is its count-only posterior mean. The full-profile posterior and Corollary~\ref{cor:profile} still apply, but the random permutation does not in general make conditioning on the other counts redundant. The empirical Good--Turing estimate, typically written $(\ell+1)K_{\ell+1}/(nK_\ell)$ when $K_\ell>0$ and $\ell<n$, is a further estimation step; none of the exact identities here justify substituting it into a universal learner-dependent bound without additional control.

\subsection{The largest coefficients valid for every learner}
For each attainable positive multiplicity, define
\begin{equation}
 \underline\theta_{n,\ell}
 =\min\{\theta_{n,\ell}(h):h\in\cH_{N,n},\ h_\ell>0\}.
 \label{eq:minimum}
\end{equation}
The minimum exists because the set of profiles is finite.

\begin{theorem}[Sharp multiplicity-only coefficients]\label{thm:sharp}
Fix an exchangeable sampling prior, a labeling prior $\cF$ independent of $D$, $N$, and $n$. Let $b_\ell$ be constants for the attainable positive multiplicities. The following statements are equivalent:
\begin{enumerate}
\item For every learner and every positive-probability dataset,
\[
 \Risk(A\mid Z)-\Opt(Z)\ge\sum_{\ell=1}^n b_\ell B_\ell(A,Z).
\]
\item For every learner,
\[
 \Risk(A)-\Opt\ge\sum_{\ell=1}^n b_\ell\E[B_\ell(A,S)].
\]
\item $b_\ell\le\underline\theta_{n,\ell}$ for every attainable positive multiplicity.
\end{enumerate}
Here the learner class is unrestricted, as in Section~\ref{sec:model}. Terms corresponding to unattainable multiplicities are zero and may be omitted.
\end{theorem}
\begin{proof}
The third assertion implies the first by Corollary~\ref{cor:profile}, since $B_\ell\ge0$; averaging gives the second. For the converse, suppose $b_\ell>\theta_{n,\ell}(h)$ at a feasible profile $h$ with $h_\ell>0$. Under the fixed labeling prior, start from a posterior-optimal predictor. When $H=h$, select uniformly one observed point with count $\ell$ and replace its known label by any different label; leave all other predictions optimal. Conditional on this profile, the excess risk is exactly $\theta_{n,\ell}(h)$ and $B_\ell=1$. Outside this profile the excess risk and all $B_j$ are zero. Thus the first assertion fails. The unconditional excess risk is $\Pp(H=h)\theta_{n,\ell}(h)$, whereas the claimed penalty is $\Pp(H=h)b_\ell$. Feasibility ensures $\Pp(H=h)>0$, so the second assertion fails too.
\end{proof}

Sharpness permits learners to concentrate errors on a selected profile; restricted algorithm or hypothesis classes may admit larger coefficients. The witness need not be permutation-equivariant for arbitrary labeling priors. For independent fair labels, however, fair predictions on unseen points and uniform selection within the multiplicity class give an equivariant witness. Thus symmetry alone does not repair the theorem.

\begin{corollary}[Validity of the count-only coefficient]\label{cor:collapse}
The bounds in Theorem~\ref{thm:sharp} hold with $b_\ell=\tau_{n,\ell}$ for all attainable positive multiplicities if and only if, for every such $\ell$, $\theta_{n,\ell}(h)$ is constant over all feasible profiles with $h_\ell>0$.
\end{corollary}
\begin{proof}
By Proposition~\ref{prop:average}, $\tau_{n,\ell}$ is a weighted average of these profile coefficients. Every feasible profile with $h_\ell>0$ has strictly positive weight, proportional to $h_\ell\Pp(H=h)$. An average is at most the minimum exactly when all its positively weighted terms are equal. Apply Theorem~\ref{thm:sharp}.
\end{proof}

For the counterexample, the singleton-containing profiles are $(1,1,1,0)$ and $(0,3,0,0)$. Their singleton coefficients are respectively \eqref{eq:counter-post} and $1/3$. Thus the sharp universal singleton coefficient is $118273/358140$, strictly below $\tau_{3,1}$. The improvement supplied by a realized profile is retained by \eqref{eq:profile-bound} and lost by taking the minimum in \eqref{eq:minimum}.

\subsection{Relation to Johnson's sufficientness postulate}\label{sec:sufficientness}
The posterior mean $q_x(c)$ is also the predictive probability that the next input equals $x$. Requiring it to depend on the sample only through $n$ and $c_x$ is a form of Johnson's sufficientness postulate. \citet{zabell1982} develops classical characterizations of Dirichlet prediction from such conditions, imposed across sample sizes with additional assumptions. See also \citet{lee2013} for the species-sampling setting.

Corollary~\ref{cor:collapse} instead fixes $N$, $n$, and the prior, and characterizes a risk bound valid for every learner using posterior means for positive counts only. Without conditions across sample sizes or for unseen categories, it implies neither a Dirichlet characterization nor uniqueness of the prior.

\paragraph{A case where count-only conditioning is sufficient.}
For comparison, if $W_i$ have a common Gamma distribution with shape $a>0$ and rate $b>0$, then $D$ has the symmetric Dirichlet distribution. Multiplication of its density by $L_c$ gives a Dirichlet posterior with parameters $(a+c_1,\ldots,a+c_N)$, so
\[
 \theta_{n,\ell}(h)=\frac{a+\ell}{Na+n}=\tau_{n,\ell}.
\]
The profile dependence disappears, as required by Corollary~\ref{cor:collapse}. This example checks the formulas and isolates a setting where the original proof strategy works. Our correction does not impose this prior or replace Feldman's finite list by Gamma weights.

\section{Consequences for the cited results}\label{sec:dependencies}
Unqualified result and equation numbers below refer to \citet{feldman2021}, arXiv version 4; references to our results are identified explicitly.

\paragraph{Disproved statements.}
Lemma 2.1, proved through Lemma A.1, identifies the full-sample posterior mass with the count-only coefficient. Our counterexample disproves that identity and both forms of Theorem 2.3. Our Theorem~\ref{thm:exponential} also rules out repairing its averaged form by a universal positive constant multiplier.

The subpopulation extension, Theorem 3.2, also fails when subpopulations are point masses on distinct domain points. The two prediction distributions in Definition 3.1 then coincide, making every learner subpopulation-coupled with coupling parameters one. The bound reduces to Theorem 2.3, disproved by our Proposition~\ref{prop:failure}. We do not analyze the coupling argument for nontrivial subpopulations.

\paragraph{Affected proof steps.}
In the noisy-label Theorem 2.4, equation (7) replaces full posterior masses by the singleton coefficient. The population-risk conclusions of Corollary 4.6 and Theorem 4.8 invoke that bound, so these conversions require separate justification. We do not supply counterexamples to each noisy-label or privacy statement or resolve their other assumptions and conventions.

\paragraph{Calculations independent of the conditioning error.}
Several intermediate calculations do not use the failed identity. The singleton-count formula and equation (8) in Section 2.5 are marginal occupancy calculations. The first inequality of Lemma 2.6 bounds the count-only coefficient itself; such a bound does not by itself bound the coefficient at every profile. Likewise, the label-prediction arguments in Lemma 4.4 and the first part of Theorem 4.8 do not rely on the frequency-conditioning step. Their conversion into excess population risk is the part requiring care. This dependency analysis does not certify every other statement in those sections.

\paragraph{Available repair and unresolved scope.}
Our Theorem~\ref{thm:decomposition} supplies an exact noiseless replacement. Under the additional fixed-$n$ condition of our Corollary~\ref{cor:collapse}, the original noiseless coefficients are valid. More generally, a uniform profile-wise lower bound on $\theta_{n,\ell}$ would suffice. The present counterexamples do not settle the proposed Zipf or other fixed long-tail conclusions; recovering them requires controlling these posterior coefficients, or the resulting learner-weighted expectation.

\section{Discussion}
Full-sample posterior frequencies give a valid lower bound in the original normalized-weight model for every independent labeling prior and every learner. The sharp-coefficient theorem identifies exactly when count-only coefficients suffice. A next step is to bound the corrected penalty for specified long-tailed priors or algorithm families. This requires controlling the posterior coefficient or its learner-weighted expectation; concentration of the normalization sum alone does not supply that control.

\section*{AI Disclosure}
Generative AI was used to assist with mathematical exploration, drafting, and LaTeX preparation.

\bibliography{references}

@misc{feldman2021,
  author = {Vitaly Feldman},
  title = {Does Learning Require Memorization? A Short Tale about a Long Tail},
  year = {2021},
  howpublished = {arXiv:1906.05271v4},
  note = {Version 4, January 10, 2021. Conference version appeared in STOC 2020},
  url = {https://arxiv.org/abs/1906.05271v4}
}

@article{good1953,
  author = {I. J. Good},
  title = {The Population Frequencies of Species and the Estimation of Population Parameters},
  journal = {Biometrika},
  volume = {40},
  number = {3--4},
  pages = {237--264},
  year = {1953},
  doi = {10.1093/biomet/40.3-4.237}
}

@article{lee2013,
 author = {Jaeyong Lee and Fernando A. Quintana and Peter M{\"u}ller and Lorenzo Trippa},
 title = {Defining Predictive Probability Functions for Species Sampling Models},
 journal = {Statistical Science},
 volume = {28},
 number = {2},
 pages = {209--222},
 year = {2013},
 doi = {10.1214/12-STS407}
}

@article{james2009,
 author = {Lancelot F. James and Antonio Lijoi and Igor Pr{\"u}nster},
 title = {Posterior Analysis for Normalized Random Measures with Independent Increments},
 journal = {Scandinavian Journal of Statistics},
 volume = {36},
 number = {1},
 pages = {76--97},
 year = {2009},
 doi = {10.1111/j.1467-9469.2008.00609.x}
}

@article{zabell1982,
 author = {S. L. Zabell},
 title = {{W. E. Johnson}'s ``Sufficientness'' Postulate},
 journal = {The Annals of Statistics},
 volume = {10},
 number = {4},
 pages = {1090--1099},
 year = {1982},
 doi = {10.1214/aos/1176345975}
}

\appendix
\section{Exact-arithmetic certificate}\label{app:certificate}
Rescale all raw weights by four, which leaves $D$ unchanged. Let $V_i=4W_i\in\{1,2\}$, with probabilities $2/3$ and $1/3$, and let $s(v)$ count the coordinates of $v\in\{1,2\}^3$ equal to two. Every expectation used in Section~\ref{sec:counterexample} can be computed from the eight-term formula
\begin{equation}
 \E[g(D)]=\sum_{v\in\{1,2\}^3}
       \frac{2^{3-s(v)}}{27}\,
       g\left(\frac{v}{v_1+v_2+v_3}\right).
 \label{eq:eight}
\end{equation}
The eight vectors and their weights are shown in Table~\ref{tab:weights}. This is equivalent to enumeration over the $27$ equally likely list-entry choices, but combines repeated entries.
\begin{table}[ht]
\centering
\caption{Exact normalized vectors in the counterexample.}\label{tab:weights}
\begin{tabular}{ccc}
\toprule
Raw vector $v$ & Probability & Normalized vector $D$\\
\midrule
$(1,1,1)$ & $8/27$ & $(1/3,1/3,1/3)$\\
$(2,1,1)$ & $4/27$ & $(1/2,1/4,1/4)$\\
$(1,2,1)$ & $4/27$ & $(1/4,1/2,1/4)$\\
$(1,1,2)$ & $4/27$ & $(1/4,1/4,1/2)$\\
$(2,2,1)$ & $2/27$ & $(2/5,2/5,1/5)$\\
$(2,1,2)$ & $2/27$ & $(2/5,1/5,2/5)$\\
$(1,2,2)$ & $2/27$ & $(1/5,2/5,2/5)$\\
$(2,2,2)$ & $1/27$ & $(1/3,1/3,1/3)$\\
\bottomrule
\end{tabular}
\end{table}
\FloatBarrier

In particular, substituting the four expressions used in the counterexample into \eqref{eq:eight} gives
\begin{align*}
\E[D_1(1-D_1)^2]&=\frac{11371}{81000},&
\E[D_1^2(1-D_1)^2]&=\frac{226313}{4860000},\\
\E[D_1D_2^2]&=\frac{5969}{162000},&
\E[D_1^2D_2^2]&=\frac{118273}{9720000}.
\end{align*}
Taking the corresponding ratios yields \eqref{eq:counter-tau} and \eqref{eq:counter-post}. Thus the finite counterexample can be checked directly from Table~\ref{tab:weights}, without external code.

\section{Leading terms for the exponential relative gap}\label{app:limits}
Let $\varepsilon=2^{-n/2}$, $u_n=(2/3)^{n-1}$, and use the definitions of $a_n,b_n,z_n,v_n$ from Theorem~\ref{thm:exponential}. Table~\ref{tab:limits} records the contribution of each raw configuration to the indicated scaled expectation, in the limit along even $n$. The first row combines the all-small and all-large configurations because they yield the same normalized vector. The other rows have probabilities $4/27$ (one large weight) or $2/27$ (two large weights), as indicated.
\begin{table}[ht]
\centering
\caption{Limits of individual contributions to the four scaled expectations.}\label{tab:limits}
\begin{tabular}{lccccc}
\toprule
Raw configuration & Probability & $a_n/\varepsilon$ & $b_n/u_n$ & $z_n/\varepsilon$ & $v_n/\varepsilon^2$\\
\midrule
All equal & $1/3$ & $0$ & $1/27$ & $0$ & $0$\\
$(1,\varepsilon,\varepsilon)$ & $4/27$ & $0$ & $0$ & $0$ & $0$\\
$(\varepsilon,1,\varepsilon)$ & $4/27$ & $4/27$ & $0$ & $4/27$ & $4/27$\\
$(\varepsilon,\varepsilon,1)$ & $4/27$ & $4/27$ & $0$ & $0$ & $0$\\
$(1,1,\varepsilon)$ & $2/27$ & $0$ & $0$ & $0$ & $1/27$\\
$(1,\varepsilon,1)$ & $2/27$ & $0$ & $0$ & $0$ & $0$\\
$(\varepsilon,1,1)$ & $2/27$ & $1/27$ & $0$ & $0$ & $0$\\
\midrule
Sum & $1$ & $1/3$ & $1/27$ & $4/27$ & $5/27$\\
\bottomrule
\end{tabular}
\end{table}
\FloatBarrier
For example, the third row contributes
\[
 \frac4{27}\frac{\varepsilon}{1+2\varepsilon}
 \left(\frac{1+\varepsilon}{1+2\varepsilon}\right)^{n-1}
\]
to $a_n$, and $(4/27)\varepsilon^2(1+2\varepsilon)^{-(n+1)}$ to $v_n$. The first row contributes $(1/27)(2/3)^{n-1}$ to $b_n$ exactly. The fifth row contributes $(2/27)(2+\varepsilon)^{-(n+1)}$ to $v_n$. The limits use $n\varepsilon\to0$, $(2/3)^n/\varepsilon\to0$, and $2^{-n}/(2/3)^n\to0$. The remaining terms either have one of these vanishing ratios or contain $\varepsilon^{n-1}$, whose contribution is smaller than every fixed exponential rate. Since the number of configurations is finite, summing the limits proves \eqref{eq:four-limits}.

\begin{table}[ht]
\centering
\caption{Finite-sample values for the family \eqref{eq:family}, computed with exact rational arithmetic and rounded for display.}\label{tab:family}
\begin{tabular}{rccc}
\toprule
$n$ & $\tau_{n,1}$ & $q_{n,1}$ & $\tau_{n,1}/q_{n,1}$\\
\midrule
8 & $0.158246$ & $0.0839837$ & $1.88424$ \\
16 & $0.0609648$ & $0.00494029$ & $12.3403$ \\
32 & $0.0235565$ & $1.90757\times10^{-5}$ & $1234.9$ \\
64 & $0.00380177$ & $2.91038\times10^{-10}$ & $1.30628\times10^{7}$ \\
\bottomrule
\end{tabular}
\end{table}

\section{Evaluating the corrected coefficients}\label{sec:integral}
The ratio in \eqref{eq:posterior} defines the corrected weights for any prior. For independently normalized positive weights it can be reduced to integrals in one variable, avoiding enumeration over all raw-weight assignments.

Suppose $W_1,\ldots,W_N$ are independent with a common law on $(0,\infty)$ and $D_i=W_i/T$. For integers $j\ge0$, $t>0$, and a profile $h=(h_0,\ldots,h_n)$, define
\[
 \phi_j(t)=\E[W^j e^{-tW}],\qquad
 F_h(t)=\prod_{j=0}^n\phi_j(t)^{h_j}.
\]
Each $\phi_j(t)$ is finite for $t>0$, even if the raw moments of $W$ are infinite.

\begin{proposition}[One-dimensional integral formula]\label{prop:integral}
For $h\in\cH_{N,n}$ and $h_\ell>0$, including $\ell=0$,
\begin{equation}
 \theta_{n,\ell}(h)
 =\frac1n
 \frac{\displaystyle\int_0^\infty t^n\phi_{\ell+1}(t)
            \phi_\ell(t)^{h_\ell-1}
            \prod_{j\ne\ell}\phi_j(t)^{h_j}\,\mathrm dt}
      {\displaystyle\int_0^\infty t^{n-1}F_h(t)\,\mathrm dt}.
 \label{eq:integral}
\end{equation}
Both integrals are finite and the denominator is positive.
\end{proposition}
\begin{proof}
For $T>0$ and an integer $m\ge1$,
\[
 T^{-m}=\frac1{\Gamma(m)}\int_0^\infty t^{m-1}e^{-tT}\,\mathrm dt.
\]
For a count vector $c$ with profile $h$, independence and Tonelli's theorem give
\[
 \E\left[\prod_iD_i^{c_i}\right]
 =\frac1{\Gamma(n)}\int_0^\infty t^{n-1}
          \prod_i\phi_{c_i}(t)\,\mathrm dt.
\]
The numerator of \eqref{eq:posterior} increases one exponent from $\ell$ to $\ell+1$ and replaces $T^{-n}$ by $T^{-(n+1)}$. Apply the same identity with $m=n+1$. The ratio of the Gamma factors is $\Gamma(n)/\Gamma(n+1)=1/n$, yielding \eqref{eq:integral}. The expectations of the corresponding products of probabilities are finite, bounded by one, and the denominator is positive. This also verifies the stated integrability without separate raw-moment assumptions.
\end{proof}

For a finite raw list, $\phi_j(t)=N^{-1}\sum_{r=1}^N\pi_r^j e^{-t\pi_r}$. Equation~\eqref{eq:integral} is a computational representation; small likelihoods may require rescaling or logarithmic evaluation. For the three-point example, quadrature gives singleton coefficients $0.330242363322$ and $0.333333333333$ for profiles $(1,1,1,0)$ and $(0,3,0,0)$, matching \eqref{eq:counter-post} and \eqref{eq:counter-distinct}. With absolute and relative quadrature tolerances $10^{-12}$, the maximum observed discrepancy over all six feasible profile--multiplicity pairs was at most $1.2\times10^{-16}$. This diagnostic is not a certified numerical-error bound. The exact values follow independently from the finite sum in Appendix~\ref{app:certificate}; the quadrature comparison is only an implementation check.

\end{document}